\documentclass[11pt]{article}

\usepackage[margin=1in]{geometry}
\usepackage{graphicx}
\usepackage{booktabs}
\usepackage{multirow}
\usepackage{amsmath}
\usepackage{microtype}
\usepackage[numbers,sort&compress]{natbib}
\usepackage{xcolor}
\usepackage[colorlinks=true,linkcolor=blue!60!black,citecolor=blue!60!black,urlcolor=blue!60!black]{hyperref}
\hypersetup{
  pdftitle={The Other Half of the Memory Wall: Serving 35B MoEs from SSD with Trained Routing Prediction},
  pdfauthor={Yu Lin, Yiming Wang, Runyuan Cai, Hanze Liu, Xiaodong Zeng},
  pdfsubject={Streaming MoE inference from SSD with prediction-as-routing},
  pdfkeywords={Mixture-of-Experts, SSD offloading, routing prediction, LoRA, quantization, on-device inference}
}
\usepackage{xurl}
\newcommand{\edge}{\textsc{Edge0}}

\title{\textbf{The Other Half of the Memory Wall:\\
Serving 35B MoEs from SSD with Trained Routing Prediction}}

\author{Yu Lin\thanks{Corresponding author. E-mail: yu.lin@autoark.ai}, Yiming Wang\thanks{yiming.wang@autoark.ai}, Runyuan Cai\thanks{runyuan.cai@autoark.ai}, Hanze Liu\thanks{hanze.liu@autoark.ai}, Xiaodong Zeng\thanks{xiaodong.zeng@autoark.ai}\\
\small AutoArk}

\date{September 2026}

\begin{document}
\maketitle

\begin{abstract}
Mixture-of-experts (MoE) inference on consumer hardware is bounded by weight
memory: a 35B-class model is 19.5\,GB at 4-bit, and sparsity shrinks the compute
per token, not the bytes that must be held. Naive offloading to SSD does not
help on its own, because layer $N{+}1$'s experts must be chosen before layer
$N$'s output exists, so the reads cannot start early enough to hide behind
compute. We present \edge{}, a streaming MoE inference engine that closes the
gap with a \emph{prerouter}: a per-layer head predicts the next layer's routing
one token ahead, and the prediction is consumed as the routing itself, so the
staged expert set equals the routed set and nothing is dropped. An unmerged
\emph{recovery LoRA}, trained on the student path, pays back the quality lost
to int4 quantization and routing replacement. On a single 24\,GB machine,
\edge{} serves a 35B MoE at ${\approx}$20\,tok/s inside 3\,GiB of peak active
memory, within a few points of its fp16 teacher on average across five public
benchmarks. An 8B tier runs on the same framework, and the framework,
checkpoints, and adapters are open source.
\end{abstract}

\section{Introduction}
\label{sec:intro}

In 1995 Wulf and McKee named the memory wall: processors improving faster than
DRAM, with machines spending an ever-larger share of their cycles waiting for
memory
\citep{wulf1995hitting}. Thirty years later the wall confronts AI inference, and
its arithmetic is unkind to the datacenter-free deployment of large models.
Decode moves a few FLOPs per byte of weight it reads while hardware ridge
points sit orders of magnitude higher; the constraint is bytes, and the bytes
live in memory.

Memory holds two kinds of data that behave very differently. \emph{State}, the KV
cache, is dynamic and grows with every generated token. \emph{Weights} are
static, fixed the day training ends. The industry attacked the dynamic half: multi-head latent attention (MLA)
compressed KV state by an order of magnitude \citep{deepseekv2}, sparse
attention capped it at a fixed horizon \citep{deepseekv32}, linear state
models removed the explicit cache altogether \citep{mamba}. The bill that
could not be renegotiated (tens of gigabytes of weights) was answered with
one move: put it in a datacenter, where expert parallelism and sharded serving
absorb it \citep{rajbhandari2022deepspeedmoe}.

The two standard local answers both fail at 35B scale. Quantization bottoms
out at 4 bits; below that, error grows until the model is unusable, and no
post-hoc technique recovers it \citep{frantar2023gptq,lin2024awq}. MoE
sparsity helps on a different axis: a 35B MoE activates about 3B parameters
per token, which shrinks what you \emph{compute}, not what you \emph{store}.
Nineteen and a half gigabytes still have to be somewhere, and on a 24\,GB desktop
shared with an operating system, that somewhere is the machine's entire
memory, crowding out the OS and everything else the user is running.

\edge{} changes where the weights live. Expert weights stay on SSD and are
mmap-streamed into memory on demand, and peak memory is bounded by the active
set instead of the parameter count. On-demand loading by itself is not fast
enough: at every decode step, layer $N{+}1$'s expert selection depends on
layer $N$'s output, so a naive streaming engine stalls on disk latency once
per layer per token. \edge{} removes the stall with a trained
\emph{prerouter}: a small per-layer head that predicts layer $N{+}1$'s routing
from layer $N$'s state at the previous token, so expert reads overlap the
forward pass. The prediction is consumed as the routing itself: the staged
expert set and the routed set are identical by construction, and the
approximation that other pre-gating schemes absorb at inference time through
fallback loads and dropped tokens is instead paid once, in training, and
recovered there.

Approximate routing and 4-bit quantization both cost quality. \edge{} trains a
\emph{recovery LoRA} on top: the int4 base is frozen, a low-rank adapter is
distilled from the fp16 teacher under the student routing path, and the
adapter is served \emph{unmerged} as a parallel delta. Merging and
re-quantizing to 4-bit erases most of the adapter's effect (\S\ref{sec:lora}).

Contributions:
\begin{enumerate}
\item \textbf{SSD as the weight tier.} An expert executor that streams MoE
expert weights from disk on demand, with routing math pinned bit-identical to
the vendored models and every executor path verified element-wise against the
dequantized reference, plus a slot mechanism that removes the per-step
stack-rebuild tax
(\S\ref{sec:streaming}).
\item \textbf{Prediction as routing.} A cross-token prerouter whose prediction
\emph{is} the routing at decode, so staged decode drops nothing and expert reads
overlap compute, worth $+80$ to $+84\%$ decode on a machine where the checkpoint
does not fit, together with the training recipe that makes a language model work
under replaced routing
(\S\ref{sec:prerouter}, \S\ref{sec:training}, \S\ref{sec:ablations}).
\item \textbf{Unmerged recovery LoRA.} Merging an adapter into an int4 base and
re-quantizing destroys most of its effect; we show that serving it as a parallel
delta avoids this at negligible cost (\S\ref{sec:lora}).
\item \textbf{Two open tiers, end to end.} A 35B and an 8B model released as
checkpoint plus adapters that load together, each within a few points of its
fp16 base on average (\S\ref{sec:eval}).
\end{enumerate}

\section{Background and Related Work}
\label{sec:related}

\paragraph{MoE routing.} Production sparse MoEs route each token to $K$ of
$E$ experts per layer
\citep{shazeer2017outrageously,lepikhin2020gshard,fedus2021switch,jiang2024mixtral}. Two routing families
dominate. Softmax-top-k (Qwen3.6): exact softmax over expert logits,
top-$k$, renormalize. Sigmoid-group (DeepSeek-V3 \citep{deepseekv3}, Ling
\citep{ling2025}): sigmoid scores; groups ranked by the sum of their top two
scores, best $G$ groups survive; top-$k$ within survivors; weights are the raw
sigmoid values, renormalized and scaled. \edge{} implements both verbatim and
reuses the same two functions for prerouter training and inference
(\S\ref{sec:prerouter}).

\paragraph{Pre-gating and the lead distance.} Pre-gated MoE
\citep{hwang2024pregated} selects the next block's experts within the same
token: the decision is produced after block $N$'s attention and consumed before
block $N{+}1$'s weights are fetched. That schedule does not survive contact
with a streaming engine, and \edge{} leads by a full token instead
(\S\ref{sec:prerouter}, where we price the per-layer
alternative).

\paragraph{KV-side compression.} MLA \citep{deepseekv2}, sparse attention
\citep{deepseekv32}, and linear-state models \citep{mamba} compress dynamic
state, and serving systems page it instead of compressing it
\citep{kwon2023pagedattention}. They are orthogonal to \edge{}, and \edge{}
benefits from them: its 8B tier is an MLA + MoE hybrid whose compressed
attention leaves more of the memory budget for the active expert set.

\paragraph{Quantization.} GPTQ/AWQ-class methods \citep{frantar2023gptq,
lin2024awq} made 4-bit the practical working point for post-training
quantization; below 4 bits, post-hoc error grows fast enough that deployed
local models rarely go lower. \edge{} uses int4
affine group-64 expert weights and accepts the loss openly, then recovers most
of it with distillation. QLoRA established that a 4-bit base carrying adapters
can approach 16-bit fine-tuning quality \citep{dettmers2023qlora}; we take the
4-bit base as given and ask what the adapter itself survives at serve time
(\S\ref{sec:lora}). Closest in approach is expert-skip
self-distillation \citep{lv2026skip}, which shows a post-trained MoE tolerates
halved expert counts when the model is trained for it; our routing-replacement
training is the same observation applied to prediction-as-routing.

\paragraph{Offloading.} Expert offloading is well
studied: llama.cpp's \texttt{--cpu-moe} \citep{llamacpp} keeps MoE experts in
CPU/system memory, PowerInfer \citep{song2023powerinfer} splits neurons into
hot and cold sets across GPU and CPU, Mixtral-offloading \citep{eliseev2023fast}
and MoE-Infinity \citep{xue2024moeinfinity} cache experts by popularity or by
reuse distance, and FlexGen \citep{sheng2023flexgen} extends the hierarchy to
disk for weights and KV together. All of them move the footprint rather than
shrinking it: the weights still occupy tens of gigabytes, and the ones that
reach disk do so without knowing which experts the
next step needs.

\section{System Design}
\label{sec:design}

\begin{figure}[t]
\centering
\includegraphics[width=\linewidth]{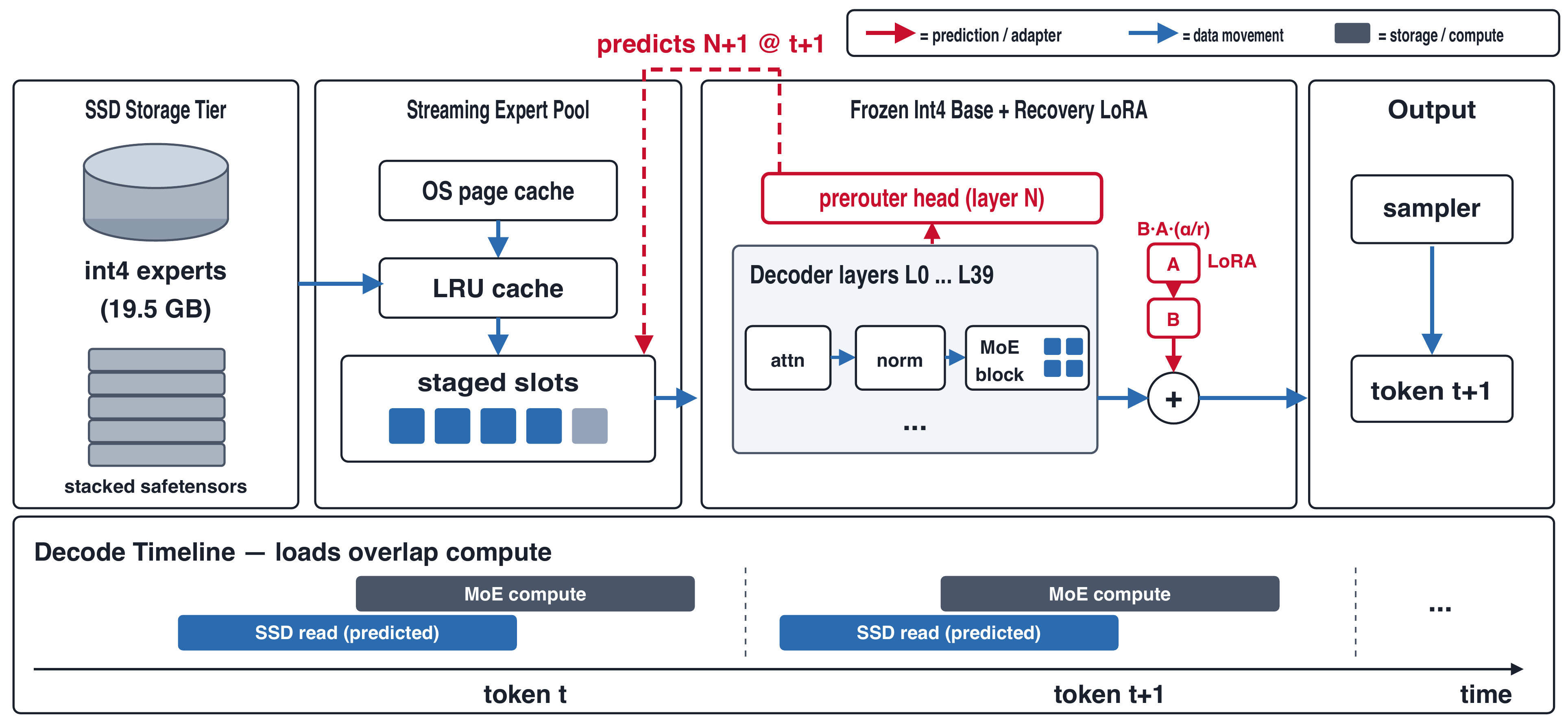}
\caption{The \edge{} system. Expert weights stay on SSD as int4 per-layer
stacked safetensors and are mmap-streamed on demand into a bounded streaming
pool (OS page cache, LRU, staged fixed-slot double buffer). A frozen int4 decoder stack computes with a parallel, unmerged
recovery LoRA branch,
$y = W_{\mathrm{int4}}(x) + \tfrac{\alpha}{r}\,BA\,x$,
combined at a sum node. The prerouter head owned by layer $N$ predicts
layer $N{+}1$'s routing one token ahead (red path, ``prediction is the
routing''), so the SSD reads that fill the staged slots overlap the forward
pass; the decode timeline at the bottom shows the predicted read of token
$t{+}1$ running concurrently with the compute of token $t$.}
\label{fig:system}
\end{figure}

\edge{} is a streaming MoE inference framework: all MLX code lives behind a
backend facade (\texttt{core}/\texttt{nn}/\texttt{io}/\texttt{quant}), and
core logic (model specs, prerouter, streaming pool, server) depends only on
that facade, so additional backends implement the same surface. Models are
described by a single \texttt{MoESpec} (expert count, $K$, routing family,
quantization, weight layout, key templates); one generic streaming layer
serves every model. LoRA and prerouter weights are safetensors files with
provenance metadata, resolved from the model directory, and never merged into
the base. Figure~\ref{fig:system} shows the system architecture and the
decode-time dataflow.

\subsection{SSD Streaming Expert Layers}
\label{sec:streaming}

\paragraph{Problem.} A 40-layer, 256-expert MoE holds 453\,MB of expert
weights per layer at int4---1.77\,MB for each of its 256 experts---or 18\,GB (16.9\,GiB) of
routed experts inside a 19.5\,GB
checkpoint whose remainder is attention, embeddings, the shared expert, and
int4 group scales. That is three quarters of the RAM of a 24\,GB machine,
before KV cache,
the operating system, and anything else the user runs. \edge{} mmaps the quantized expert safetensors
and reads byte ranges on demand; the operating-system page cache carries hot
data and peak memory tracks the \emph{active} set, not the parameter count.

\paragraph{Executor paths.} Four paths, from the correctness baseline to the prefill bulk path:
\begin{enumerate}
\item \emph{exact}: deduplicated on-demand bundle build, stack, quantized
gather. The correctness baseline for every other path.
\item \emph{staged}: fixed-slot double buffering, the decode workhorse.
Routing indices are mapped through a slot table with \texttt{take}; indices
never leave the GPU. Repeated expert sets reuse cached graph nodes; persistent
sticky-slot tensors are updated in place (\texttt{incr\_stack}) so a step
rewrites only the experts that changed instead of rebuilding the layer's nine
stacked tensors---gate, up, and down weights, each with its scales and
biases---across all 40 layers.
\item \emph{hot}: a fixed set of LRU-resident hot experts per layer, the same
hot/cold
split as PowerInfer and the popularity caches of Mixtral-offloading and
MoE-Infinity \citep{song2023powerinfer,eliseev2023fast,xue2024moeinfinity};
hits take the stacked gather, misses fall to exact.
\item \emph{whole-layer}: all experts of a layer loaded in one shot for
prefill, which the checkpoint's per-layer stacked layout makes nine direct
reads---the same three quantized projections---rather than $256\times9$ builds;
CPU load overlaps the previous layer's
GPU execution.
\end{enumerate}

\paragraph{Math contract.} Every path computes
$\mathrm{down}(\mathrm{silu}(\mathrm{gate}(x))\cdot \mathrm{up}(x))$ with the
same quantized gather kernel, and the test suite compares each path
element-wise against the dequantized reference (relative L2 $<1\%$; measured
residual $\approx0.24\%$ is the bf16-internal precision of the kernel
itself).
This contract is what allows the executor to switch paths freely per layer,
per phase, without changing outputs.

\paragraph{Why offload can win outright.} A fully-resident engine on this
hardware is not the fast baseline it appears to be, because the model does not
fit: the Mac mini M4 Pro has 24\,GB, and a vanilla mlx-lm server with all
19.5\,GB of 4-bit weights resident decodes at 3.9\,tok/s occupying 18.2\,GiB,
while \edge{}'s $K{=}4$ profile decodes at 20.4\,tok/s occupying
2.9\,GiB.
\footnote{Same machine, same decode protocol, think-mode on.} The
resident path holds 18.2\,GiB that cannot be reclaimed, which leaves almost
nothing of the 24\,GB for the KV cache and the operating system; \edge{} pays
instead in graph building and tensor assembly, a cost of our streaming
engine rather than of the weights, and the subject of \S\ref{sec:limits}. The oracle
experiment makes that cost explicit: with perfect routing prediction and an
unbounded cache, the only remaining cost is per-step tensor assembly, which
\texttt{incr\_stack} then attacks directly (Appendix~\ref{app:assembly}).

\subsection{The Prerouter: Prediction Is the Routing}
\label{sec:prerouter}

\paragraph{Motivation.} Expert selection for layer $N{+}1$ depends on layer
$N$'s output, which does not exist yet when layer $N{+}1$'s loads would need
to start. Waiting serializes disk latency into every step. The prerouter
breaks the dependency with a double shift: the head owned by layer $N$ runs
at token $t$ on layer $N$'s post-attention norm output and predicts layer
$N{+}1$'s routing at token $t{+}1$. Layer $N{+}1$ consumes the prediction
made one token earlier; the SSD read that fills its slots overlaps the
current forward pass (Figure~\ref{fig:system}).

The lead has to be a full token, not one layer. Choosing layer $N{+}1$'s
experts right after layer $N$'s attention, as Pre-gated MoE does
\citep{hwang2024pregated}, needs a per-layer synchronization and head evaluation
that drain the GPU pipeline ($30$--$100$\,ms per step in our engine, more than
the load time it can hide, and every same-token variant we measured fell below
a plain LRU baseline), and the next layer's attention has not been computed when its
experts must be chosen. One token of lead moves the head evaluation off the
per-layer critical path: a single flush per step predicts every staged layer
at once (32 on the 35B tier, 16 on the 8B tier), and the window it opens has to
span a whole decode step, because the reads it hides are a step's worth of disk
traffic: at $K{=}4$ a layer's experts are ${\approx}7$\,MB, and the prerouter
arm still spends 101.9\,ms per step waiting on cold loads
(\S\ref{sec:ablations}).

\paragraph{Head.} Per layer:
$\mathrm{fc}_1 \to \mathrm{erf\text{-}gelu} \to \mathrm{fc}_2$, plus a
linear residual path $\ell$ that the training script warm-starts from the
\emph{next layer's} router weight (its default is zero), so training begins
from ``apply the next router directly to this hidden state'' and the MLP
learns the correction.
The input feature concatenates the hidden state with two top-$k$ one-hots:
the experts this layer actually routed to at this token, and those at the
previous token ($\mathrm{dim} = d_{\text{model}} + 2E$; for the 35B tier,
$2048 + 2\times256 = 2560$, hidden width 512). Heads are fp16, the training
export precision; running them in fp32 costs measurable time for no accuracy
benefit. The 35B tier carries 33 heads (owners 6--38) and the 8B tier 16
(owners 7--22); predictions are consumed at layers 7--38 on the 35B tier and
8--23 on the 8B tier, so 32 of 40 and 16 of 24 layers stream from a prediction
rather than from their own gate (the 35B head owned by layer 38 predicts layer
39's routing, which is never staged, so it ships without a consumer). The 8B input is $1536 + 2\times128$.

\paragraph{Prediction-as-routing.} At decode, MoE layers
route by the prerouter's logits instead of the router's, through the same
softmax-topk or sigmoid-group math as the original router
(Appendix~\ref{app:routing}). Because the
routed set \emph{is} the predicted set, the staged slots map exactly and
there is nothing to drop: the coverage-vs-quality trade-off that pre-gated
systems handle at runtime \citep{hwang2024pregated} is eliminated by
construction. What the
approximation costs is transferred to training, where it can be paid once
(\S\ref{sec:training}).

\paragraph{Feature drift.} A head's input includes the routing its layer
actually executed. At training that is the base router's selection; at decode
it is the head's own prediction, because the prediction replaced the router.
The heads are not retrained for that shift. The recovery LoRA is trained on
the student path with prerouting in place, so it sees the deployed input
distribution, and the quality measurements of \S\ref{sec:eval} price the
shift together with int4 and the routing approximation.

\paragraph{Two consumption profiles.} Both tiers run prediction-as-routing
staged decode: the 8B tier at $K{=}8$ with eight staged slots per layer, the
35B
tier at $K{=}4$ with the incremental sticky-slot stack
(\S\ref{sec:streaming}). Prefill takes the
whole-layer path on both tiers, where every expert of a layer is hit and there
is nothing to predict, and the prerouter is exercised at decode. The same
trained heads and the same stager abstraction serve both profiles; the
framework exposes them as layer options.

\subsection{Recovery LoRA: Unmerged by Design}
\label{sec:lora}

The served model is (int4 base) $+$ (routing replacement) $+$ (LoRA), and the LoRA \citep{hu2022lora} exists to recover the first two.
The sidecar is the same low-rank construction we used for multi-task and
privacy-preserving serving \citep{wang2023multilora,wang2023privatelora}, and
the same frozen-base-plus-sidecar pattern recently applied to generative-vision
personalization \citep{cai2026tinyengram}. It is trained on the student path
(routing by prerouter) with cross-entropy against the teacher's data, so the
adapter compensates quantization and routing approximation jointly.

Deployment keeps it unmerged: $y = W_{4\text{bit}}(x) +
\frac{\alpha}{r}\,BAx$ computed as a parallel delta, the
4-bit base bytes untouched. The alternative, merging the delta into the
dequantized weight and re-quantizing to 4-bit, fails on arithmetic rather than
implementation: LoRA deltas (RMS $10^{-3}$) sit below the 4-bit group step,
so requantization erases most of the weight-level delta
(34\% of the effect survives on an attention projection, 2\% on a dense
projection), and at the logits level 18\% survives.\footnote{Retention at the
logits level is measured as $1 - \lVert U{-}M\rVert / \lVert U{-}B\rVert$,
which is $0.18$ for the merged model, where $U$ = unmerged, $M$ = merged,
$B$ = base, first-token logits on a fixed prompt; weight-level retention is
$\Sigma(\text{actual}\cdot\delta)/\Sigma \delta^2$ per target.} The unmerged path costs 42\,MB of adapter weights and no measurable decode
time, and it is strictly more faithful. Adapters therefore ship as files
beside the checkpoint: one read-only base serves every adapter generation,
and retraining a tier means swapping two files.

\section{Training}
\label{sec:training}

The recipe has three phases, and all of them run on the dequantized bf16
reconstruction of the 4-bit deployment checkpoint (we train on what is
served), with the base frozen throughout.

\paragraph{Phase 1: distill the heads.} The prerouter heads are the only
trained parameters. The loss imitates the next layer's true router, so the
heads learn to predict routing rather than to fit text.

\paragraph{Phase 2: SFT on the student path.} The LoRA is attached to
attention, linear-attention, and shared-expert projections, but not to routed
experts, whose weights stream and must stay replaceable, and training runs
the full forward with prerouter routing active (the ``student path'') over
roughly two million rows of teacher-generated text. This is
the phase that makes the approximation usable: in our runs, distillation-only
checkpoints with student routing produce repetitive, collapsed text, while the
same heads plus SFT produce coherent output at identical speed and memory. The
order was not negotiable in those runs: heads first (the SFT signal otherwise
drowns the tiny head gradients), SFT second, on-policy distillation last. Chasing router agreement
harder is the wrong
objective: cross-token prediction from the previous token's hidden state is
information-limited, and what matters is whether the \emph{language model}
produces good text under the student routing.

\paragraph{Phase 3: on-policy distillation.} Phase 2 trains on text the teacher
wrote; Phase 3 trains on the student's own generations. The Phase-2 checkpoint
generates, the original fp16 base scores those tokens as teacher, and the
gradient is taken through the served path on the same trainable surface as
Phase 2. The objective is reverse KL (mode-seeking, so the student is never
asked to cover the teacher's entire support
\citep{gu2024minillm,agarwal2023gkd}), applied to the teacher's top-$k$ tokens,
with the mass outside the top-$k$ set carried by a tail term rather than
dropped \citep{huang2026taopd,dasgupta2026tail}. Phase 3 converges on one tenth
of the SFT corpus (roughly 200k rows against the $\approx$2M of Phase 2).

\paragraph{Iteration economics and the released width.} Retraining a tier for
a different routing width means swapping adapter files and nothing else. That
is what let us release the 35B tier at $K{=}4$ rather than the base model's
$K{=}8$: on the 16\,GB machine of \S\ref{sec:ablations}, in a same-session
A/B at one cache budget with the same weights at both widths, narrowing from
$K{=}8$ to $K{=}4$ nearly doubles decode ($3.3$ to $6.4$\,tok/s,
Table~\ref{tab:anticip}) and lowers peak active memory,
while the retrained tier holds the quality of
Table~\ref{tab:quality}. Width is the one knob here that moves speed and
memory together, and the frozen base is what makes turning it a file swap
rather than a training campaign.

\section{Evaluation}
\label{sec:eval}

\subsection{Setup}
\label{sec:setup}

\begin{table}[t]
\centering
\caption{The two public release tiers, with the released checkpoint and
adapters on a Mac mini M4 Pro 24\,GB. Decode and memory are the latest paired
measurements (each arm in its own process, warm, arm order rotated, medians over
3 rounds); memory is the MLX allocator peak at short contexts, since expert
weights stream from SSD and only the decoder's expert cache is resident.
Checkpoint on disk is the int4 base; the adapter and prerouter heads add
0.2\,GB. The 35B row is the $K{=}4$ production profile of
\S\ref{sec:prerouter}; the 8B row is its low-memory release profile, with a
shared LRU of 64 expert bundles (${\approx}1.5$\,GiB) separate from the
per-layer staged slots; enlarging that LRU to 1024 bundles costs
${\approx}0.9$\,GiB of the budget and lifts decode to 31.8\,tok/s. Prefill warm is measured within one process on
a 3.1k-token prompt, cold is the first request after process start. Disabling
the prerouter on the 35B tier measures 19.9\,tok/s.}
\label{tab:tiers}
\small
\begin{tabular}{lcc}
\toprule
 & \texttt{edge0-35b} & \texttt{edge0-8b} \\
\midrule
Base model & Qwen3.6-35B-A3B \citep{qwen36moe} & Ling 3.0 tiny (MLA+MoE) \citep{ling3tiny,ling2025} \\
Layers $\times$ experts & $40 \times 256$ + shared & $24 \times 128$ + shared \\
Active params / token & $\approx$3B & $\approx$1.2B \\
Routing width $K$ & 4 & 8 \\
Routing family & softmax-topk & sigmoid-group ($8{\times}4$, $\times2.5$) \\
Quantization & int4 affine g64 & int4 affine g64 \\
Checkpoint on disk & 19.5\,GB & 4.5\,GB \\
Prerouter heads & 33 & 16 \\
LoRA & $r{=}16,\ \alpha{=}32$ & $r{=}16,\ \alpha{=}32$ \\
\midrule
Decode (tok/s) & 20.4 & 28.0 \\
Prefill cold / warm (tok/s) & 113 / 140 & 500 / 1102 \\
Peak active memory & 2.9\,GiB & 1.5\,GiB \\
\bottomrule
\end{tabular}
\end{table}

Quality runs use OpenCompass on a compute server under identical
settings for \edge{} (int4 + adapters + prerouter routing) and the original
fp16 bases; they involve no timing. All throughput and memory measurements
are single-device: tier-profile numbers (Table~\ref{tab:tiers}) on a
Mac mini M4 Pro 24\,GB, the prerouter A/B (Fig.~\ref{fig:gaink}) on a MacBook
M2 with 16\,GB of
unified memory holding the release directory (18.4\,GiB: a 19.5\,GB int4
base plus 0.2\,GB of adapter and prerouter heads), a machine on which the
weights do not fit, so every step faults experts back in from the SSD, with mlx
0.30.4--0.30.6 spanning the campaigns \citep{mlx}.
Speed comparisons use same-session alternating A/B with page-cache warmup and
matched cache budgets on both arms: each arm runs in its own process, the arm
order rotates every round, both arms replay the same sampled token sequence,
and every number we report is a median over repeated runs. Single-shot
benchmarks on this hardware carry $\pm40\%$ run-to-run spread, and up to
$2.3\times$ across sessions on the 16\,GB machine, which is why no
cross-session number is used as evidence anywhere in this section.

\subsection{Quality}
\label{sec:quality}

\begin{figure}[t]
\centering
\includegraphics[width=\linewidth]{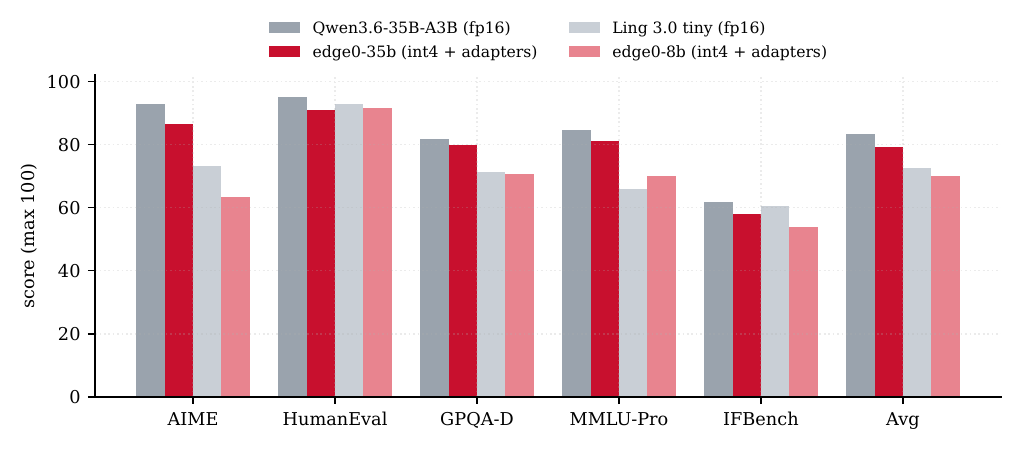}
\caption{\edge{} (int4 + prerouter routing + recovery LoRA) vs fp16 base
models, OpenCompass, identical settings. Mean per-benchmark gap 3.9 (35b)
and 2.8 (8b) points.}
\label{fig:quality}
\end{figure}

\begin{table}[t]
\centering
\caption{Quality vs fp16 base models (max 100, OpenCompass).}
\label{tab:quality}
\small
\begin{tabular}{lcccc}
\toprule
Benchmark & \texttt{edge0-35b} (int4) & Qwen3.6 (fp16) & \texttt{edge0-8b} (int4) & Ling 3.0 tiny (fp16) \\
\midrule
AIME 2026        & 86.6 & 92.7 & 63.3 & 73.3 \\
HumanEval        & 90.9 & 95.1 & 91.5 & 92.7 \\
GPQA-Diamond     & 79.8 & 81.8 & 70.7 & 71.2 \\
MMLU-Pro         & 81.0 & 84.6 & 70.1 & 65.8 \\
IFBench          & 57.9 & 61.7 & 53.9 & 60.6 \\
\midrule
Average          & 79.2 & 83.2 & 69.9 & 72.7 \\
\bottomrule
\end{tabular}
\end{table}

Table~\ref{tab:quality} and Figure~\ref{fig:quality} give the picture. The
pipeline recovers most of the joint int4-plus-routing-re\-place\-ment loss:
mean per-benchmark gaps of 3.9 points on the 35B tier and 2.8 on the 8B,
close enough that we
treat the two served tiers as quality-matched to their fp16 bases.

\subsection{The Advantage of the Prerouter}
\label{sec:ablations}

The experiments in this section run on a MacBook M2 with 16\,GB of unified
memory serving an 18.4\,GiB checkpoint, on a machine where the weights do
not fit. No artificial memory limit is imposed (no \texttt{mlock}, no wired pages,
no cgroup cap, no page-cache purge), so the only constraint is physical memory
itself and every configuration here faults experts back in from the SSD. What
must fit for the engine to run \emph{at all} is the unreclaimable MLX
allocation, and the prerouter raises it: 1.72, 1.73, and 1.79\,GiB for
on-demand streaming against 2.33, 2.60, and 3.22\,GiB ($K{=}2,4,8$). Process RSS---which counts the
file-backed expert pages as well as the allocator---peaks at 5.29, 6.12, and
6.18\,GiB for on-demand streaming against 4.91, 5.61, and 6.39\,GiB with the
prerouter. The rest of the machine is page
cache, which is reclaimable and gets displaced by residency.

\begin{figure}[t]
\centering
\includegraphics[width=\linewidth]{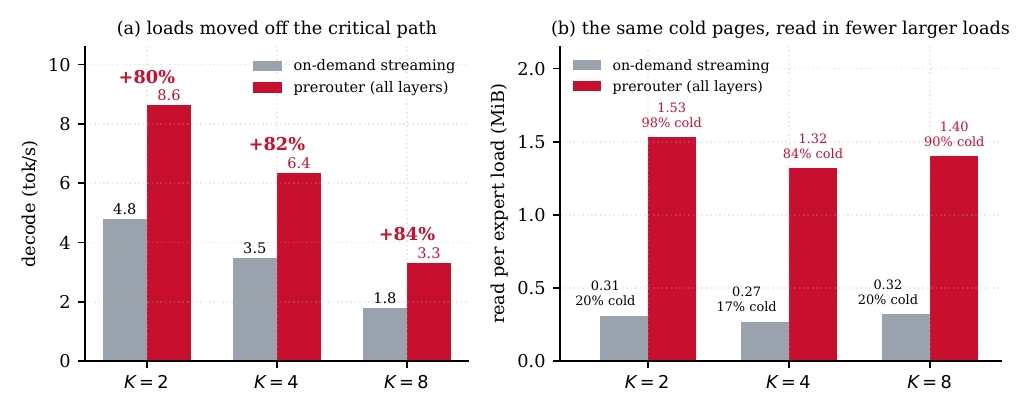}
\caption{The advantage of the prerouter, on a 16\,GB MacBook M2 with an
18.4\,GiB checkpoint that does not fit. (a) Decode throughput: pure on-demand
streaming against every staged layer prefetching its next token's experts.
(b) The same cold pages read in fewer, larger loads, with the cold
fraction of each load printed above its bar; per-load cost follows
$1.17\,\mathrm{ms} + 1.33\,\mathrm{ms}\times$cold. Same-session rotated A/B,
both arms replaying the same sampled token sequence, 3-round medians, 3/3
rounds agreeing.}
\label{fig:gaink}
\end{figure}

\begin{table}[t]
\centering
\caption{The advantage, on the same machine and in the same session. Decode
throughput with on-demand streaming against every staged layer prefetching
one token ahead.}
\label{tab:anticip}
\small
\begin{tabular}{lccc}
\toprule
$K$ & on-demand & prerouter & gain \\
\midrule
2 & 4.8 & \textbf{8.6} & $+80\%$ \\
4 & 3.5 & \textbf{6.4} & $+82\%$ \\
8 & 1.8 & \textbf{3.3} & $+84\%$ \\
\bottomrule
\end{tabular}
\end{table}

\begin{table}[t]
\centering
\caption{What the price is made of (same session, same arms). Both arms move
within a few percent of the same bytes at $K{=}2$ and $K{=}8$ (29.5 against
30.4 and 125.1 against 126.9\,MiB, prerouter against on-demand); at $K{=}4$ the
prerouter arm reads 16\% more (58.9 against 50.9\,MiB). The prerouter reads
them in fewer, larger, colder loads.}
\label{tab:granularity}
\small
\begin{tabular}{llrrrrr}
\toprule
$K$ & arm & loads/step & MiB/step & MiB/load & cold & ms/load \\
\midrule
2 & on-demand & 99.3 & 30.4 & 0.31 & 20\% & 1.56 \\
2 & prerouter & 19.3 & 29.5 & \textbf{1.53} & \textbf{98\%} & 2.41 \\
\cmidrule(lr){1-7}
4 & on-demand & 189.4 & 50.9 & 0.27 & 17\% & 1.29 \\
4 & prerouter & 44.7 & 58.9 & \textbf{1.32} & \textbf{84\%} & 2.28 \\
\cmidrule(lr){1-7}
8 & on-demand & 398.6 & 126.9 & 0.32 & 20\% & 1.44 \\
8 & prerouter & 89.5 & 125.1 & \textbf{1.40} & \textbf{90\%} & 2.36 \\
\bottomrule
\end{tabular}
\end{table}

\paragraph{The advantage is the load time moved off the critical path.}
With every staged layer prefetching, the prerouter issues the same reads
earlier---16\% more bytes at $K{=}4$
(Table~\ref{tab:granularity}). On this machine the time the main thread spends
blocked on expert loads falls from
154.9 to 46.5\,ms per step at $K{=}2$, from 244.0 to 101.9\,ms at $K{=}4$, and
from 575.0 to 211.6\,ms at $K{=}8$. The blocked time is summed over layers,
and different layers' loads overlap, so at $K{=}8$ the on-demand sum (575.0\,ms)
is slightly larger than the step it belongs to (559.1\,ms). No resource is
saturated while this
happens: the disk read is at most 12\% of the step even at its widest, the
process uses about one core of eight, and the GPU runs at 35--41\%. The cost
being removed is serialized load latency, and removing it is what the predictor
buys (Figure~\ref{fig:gaink}a).

\paragraph{Conservation, not prediction quality.} Both arms read within a few
percent of the same bytes per step at $K{=}8$ (125.1 against 126.9\,MiB) and at
$K{=}2$ (29.5 against 30.4\,MiB); only at $K{=}4$ does the prerouter arm read
more, 58.9 against 50.9\,MiB, or 16\% (Table~\ref{tab:granularity}). Moving a read off the critical path can
hide it, but it cannot delete it: the advantage comes from the cold-read time
that is exposed on that path today. With every staged layer prefetching,
decode reaches 8.6, 6.4, and
3.3\,tok/s against 4.8, 3.5, and 1.8\,tok/s on demand: $+80\%$, $+82\%$, and
$+84\%$ (Table~\ref{tab:anticip}). How much there is to win is a property of
the storage tier, not of the head: the head only has to supply the right
set.

\paragraph{Realizing it takes reuse.} Adjacent tokens agree on only about a
quarter of a layer's expert set in our traces, so most prefetched experts are
never read again, and a load that is issued and not used buys nothing. The
storage tier sets how much is available; reuse sets how much of it the
prediction collects.

\paragraph{The price is page cache, and cold pages are a granularity problem.}
Prefetching requires the predicted experts to be resident before they are
needed, and resident MLX memory is not reclaimable: at $K{=}8$ the
prerouter arm holds 1.43\,GiB more of it while the page cache loses
1.15\,GiB, four fifths of what the residency takes. The same cold pages then arrive in fewer, larger
loads: 1.40\,MiB per load at 90\% cold, against 0.32\,MiB per load at 20\%
cold for on-demand streaming (Figure~\ref{fig:gaink}b,
Table~\ref{tab:granularity}). The cost of a load is
$1.17\,\mathrm{ms} + 1.33\,\mathrm{ms}\times(\text{cold fraction})$, fitted on
the $K{=}8$ pair and reproducing every measured per-load cost within
0.13\,ms (the prerouter points within 0.07\,ms). Prefetching is cheap per
load and expensive per byte moved early.

\paragraph{Two levers not yet pulled.} Since the price is unreclaimable
residency, the cheapest remaining gain would be to stop paying for memory the
loader does not need: the stager in these runs keeps both a bundle and a
stacked tensor view of the same weights, and dropping one copy would return
about 0.45\,GiB at $K{=}8$. The second is the fill itself: with incremental
stacking it runs synchronously on the main thread here, which at $K{=}8$ costs
62.5\,ms per step, so moving it off that thread would remove a cost that never
needed to be waited for. Neither change is in the configuration measured
above.

\paragraph{One mechanism, not two.} The prerouter supplies staged decode with a
set that is correct \emph{by definition} (it is the routing), so the staged
slots map exactly and nothing is dropped. Staging alone is not a deployable
configuration: without a correct set source it either drops experts and the
output degrades, or pins a hot set that churns. Prediction chooses, staging
loads.

\subsection{Memory: What the Machine Actually Pays}
\label{sec:memory}

Peak active memory is 2.9\,GiB for the 35B tier against a 19.5\,GB checkpoint
(Table~\ref{tab:tiers}); the remainder is disk. The same weights fully
resident need 18.2\,GiB and decode at 3.9\,tok/s: 18.2\,GiB of
unreclaimable weights leave too little of the 24\,GB for the KV cache and the
operating system, so the system pages. \edge{} occupies one-sixth to
one-seventh of what the resident server holds and decodes at 20.4\,tok/s
(Table~\ref{tab:tiers}), five times faster.

Prefill behaves as the mirror image: cold first-request prefill pays the SSD
fault-ins (35B tier: 113/140\,tok/s cold/warm on a 3.1k-token prompt; 8B
tier: 500/1102), and whole-layer loading plus the page cache makes warm
prefill compute-bound. Decode, not prefill, is the constrained phase,
and every mechanism here targets it.

\section{Limitations}
\label{sec:limits}

\edge{} serves one request at a time, FIFO-serialized. Concurrency belongs
to a serving layer, not the engine, and batching changes the expert working
set in ways our per-request profiles do not model.

The streaming engine's decode is CPU-side, not storage-side: 44\,ms per step
of graph building in the 40-layer forward is the floor, and no storage-side
optimization moves it. Closing that gap needs kernel-level graph amortization
or a smaller model (Appendix~\ref{app:assembly}).

Prerouter gains scale with the waiting they remove: they shrink on
hot caches and fast storage, and they are bounded by reuse: adjacent tokens
agree on only about a quarter of a layer's experts, so a prefetched expert is
often read once and paid for twice. Its price is unreclaimable residency taken
from the page cache, which is exactly what makes the mechanism a win where
memory is tight and the storage tier is slow, the regime \edge{} was built
for.

Quality loss concentrates in long-chain reasoning: 6.1 points on AIME for
the 35B tier and 10.0 for the 8B tier; every other 8B benchmark is within 6.7
points, and MMLU-Pro is 4.3 points in the student's favour. The recovery LoRA recovers most of the pipeline's loss
everywhere else, and reasoning is where int4 plus routing replacement
remains visible.

The MLX backend is the one implementation; the backend facade is the
abstraction, and the CUDA slot is architecture, not code.

\section{Conclusion}

The weights half of the memory wall is a storage
placement decision, and everyone made it the same way. \edge{} makes the other
decision: experts live on SSD, a trained prerouter predicts routing one
token ahead so loads hide under compute, prediction is the routing so
nothing is dropped, and a distilled, unmerged LoRA pays the quality bill at
4-bit. The result is a 35B-class MoE served from a 24\,GB consumer desktop
at 20\,tok/s inside 3\,GiB of memory, an 8B hybrid at 28\,tok/s inside
1.5\,GiB,
both within a few points of their fp16 teachers, with every component (framework,
checkpoints, adapters) open. The machine on your desk is already big enough;
the weights just needed somewhere to live.

\paragraph{Artifacts.}
\begingroup\small\sloppy
Framework: \url{https://github.com/Edge0-AI/Edge0}.
Models: \url{https://huggingface.co/Edge0/Edge0-35B-A3B-preview} and
\url{https://huggingface.co/Edge0/Edge0-8B-A1B-preview}. Apache-2.0.
\endgroup

\bibliographystyle{plainnat}
\bibliography{refs}

\appendix

\section{Routing Math}
\label{app:routing}

Both functions are extracted verbatim from the vendored model
implementations and shared by resident, streaming, and prerouter paths;
parity tests pin them bit-identical.

\paragraph{Softmax-topk ($K$, renormalize).}
\begin{align*}
g &= \mathrm{softmax}(\ell) \qquad
\mathrm{inds} = \mathrm{top\text{-}k}(g,\ K)\\
w &= g[\mathrm{inds}] \,\big/\, \textstyle\sum_{\mathrm{inds}} g
\end{align*}

\paragraph{Sigmoid-group ($K$, $G$ of $n$ groups, scale $s$).}
\begin{align*}
\sigma &= \mathrm{sigmoid}(\ell)\\
\text{group score}_i &= \text{top-2 sum of } \sigma \text{ in group } i, \quad
\text{keep top } G \text{ groups, mask others to } {-\infty}\\
\mathrm{inds} &= \mathrm{top\text{-}k}(\sigma \text{ within survivors},\ K)\\
w &= s \cdot \sigma[\mathrm{inds}] \,\big/\, (\textstyle\sum \sigma[\mathrm{inds}] + 10^{-20})
\end{align*}

Selection uses the biased score; weights use the raw sigmoid, the
DeepSeek-style distinction \citep{deepseekv3}. The 8B tier runs $n{=}8$,
$G{=}4$, $s{=}2.5$, $K{=}8$.

\section{The Assembly Tax and Incremental Stacks}
\label{app:assembly}

The oracle bound (\S\ref{sec:streaming}) locates the residual gap in per-step
tensor assembly. In a separate A/B against the same pipeline without \texttt{incr\_stack},
replacing per-step stacks with persistent sticky-slot tensors updated in place
contributes a further $+34\%$ decode. In a second same-session A/B on the
$K{=}8$ configuration---native routing against the complete shipped
pipeline---decode rises from 6.8 to 12.5\,tok/s. Both pairs come from the
earlier campaign rather than the same-session sweep of
\S\ref{sec:ablations}, and from a different configuration, so their rates are
not comparable with Table~\ref{tab:anticip}; they price assembly, not
prefetching: with \texttt{incr\_stack} the pipeline rewrites only the slots
that changed, and what the prerouter does \emph{not} remove is the per-step
assembly that remains.

\end{document}